# Widening the Dialogue Workflow Modeling Bottleneck in Ontology-Based Personal Assistants

**Michael Wessel, Edgar Kalns, Girish Acharya, Andreas Kathol**

SRI International, 333 Ravenswood Avenue, Menlo Park, CA 94025, USA

firstname.lastname@sri.com

## Abstract

We present a new approach to *dialogue specification for Virtual Personal Assistants (VPAs)* based on so-called *dialogue workflow graphs,* with several demonstrated advantages over current ontology-based methods. Our new *dialogue specification language (DSL)* enables customers to more easily participate in the VPA modeling process due to a user-friendly modeling framework. Resulting models are also significantly more compact. VPAs can be developed much more rapidly. The DSL is a new modeling layer on top of our ontology-based Dialogue Management (DM) framework OntoVPA. We explain the rationale and benefits behind the new language and support our claims with concrete reduced Level-of-Effort (LOE) numbers from two recent OntoVPA projects.

## Introduction & Motivation

In 2020, *Virtual Personal Assistants (VPAs)* have become commonplace on our smartphones and in smart speakers and are expected to become even more important in 2021[1]. The VPA landscape ranges from simple "one-shot request-response" systems to control appliances and lights in an IoT home automation network, over more sophisticated task-oriented virtual specialists á la KASISTO[2] that help with complex tasks such as online banking, to freeform conversational chatbots that aim at passing the Turing Test.

With growing demands and user's expectations regarding the "intelligence" and "human-likeness" of VPAs, there is clearly a need for DM systems that go beyond the simple "one-shot request-response" model. In 2020, users expect systems that are capable of supporting a) elements of freeform chat with contextual memory (e.g., learn about the user and current situation), b) are capable of supporting complex workflows and multiple turns, and c) can combine both in a non-rigid, non-linear, naturally flowing conversation that doesn't feel like a telephone hotline support system.

We have embraced and addressed some of these challenges in our OntoVPA framework (Wessel et. al. 2018).

Most VPAs rely on one central component - the Dialogue Manager. The Dialogue Manager is typically concerned with *Dialogue Management (DM),* which encompasses *Dialogue State Tracking* (Williams, Raux, and Henderson 2016), and it also implements the *Dialogue Policy*: the computation of the system response ("system turn") based on the current state of the world and dialogue / discourse (including the previous and current "user turns").

A commonly used *conceptual* DM model is the *Dialogue Flow Graph.* We are referring to this graph as a *conceptual* model because the actual dialogue model representation might be different from a graph (e.g., in case of a rule-based manager). Still, the flow graph is a useful conceptual notion of high utility for developers and customers, given its comprehensibility. Unsurprisingly, various commercial solutions for "visual drag & drop" chatbot construction exist.

In the following, we prefer the term *Dialogue Workflow Graph (DWG)* because it emphasizes that this graph may also contain elements ("nodes and edges") of *workflow management*, i.e., elements that describe the workflow actions that need to be carried out behind the scenes of the dialogue.

In this paper, we are presenting a novel DWG Specification Language*, Dialogue Specification Language (DSL)* for short, and show that is has potential to significantly reduce the VPA modeling effort. Our DSL is also *a Domain Specific Language* in the conventional sense. It is built on top of OntoVPA.

OntoVPA is a declarative, knowledge-based system, in which a new VPA domain can be implemented with very little to no conventional programming required. OntoVPA is highly expressive and comes with built-in solutions to standard DM problems such as state tracking, anaphora resolution, contextual intent slot-filling, intent management, stack-based sub-dialogue management, etc. The rules that implement these common DM capabilities are generic and cross-domain. They are typically expensive to implement from scratch. As we will illustrate below, by relying on expressive ontology reasoning and forms of higher-order quantification, these rules can *succinctly cover large regions in the DM problem space*, hence greatly reducing the modeling effort of generic DM capabilities.

*Unfortunately*, a *much bigger LOE* is required to model the dialogue workflows by means of rules. These are obviously VPA domain-specific and hence *cannot* be transferred cross-domain easily. From our experience in developing 5 VPAs with OntoVPA, this LOE can easily account for more than 70% of the overall LOE.

Our DSL has shown great potential to reduce and simply this LOE. It also has the benefit of being more intuitive and much less technically involved: unlike plain OntoVPA, developers no longer need to use OWL and SPARQL exclusively. *Visual* workflow dialogue graphs can be generated automatically from the textual DSL specifications, which facilitates customer participation, transparency, and comprehensibility. *The DSL hence has the potential to significantly reduce the modeling effort, thus widening the dialogue workflow modeling bottleneck.*

The remainder of this paper is structured as follows. We first describe the OntoVPA framework, providing the basis for this work. The essence of ontology-based DM is illustrated by means of typical DM problems in the (ubiquitous exemplary) "Restaurant Recommendation" domain (Henderson, Thomson, and Williams 2014). We subsequently describe three different workflow representation options, also corresponding to layers building on top of each other. The lowest layer (Level 1) corresponds to plain OntoVPA dialogue workflow modeling, and the final (Level 3) abstraction layer corresponds to new DSL. We then quantitatively demonstrate the benefits of the DSL in terms of reduced modeling LOE. We conclude with a discussion of related work and a summary and outlook for future work.

## Ontology-Based Dialogue Management

The original OntoVPA is described elsewhere in full detail (Wessel 2018). It realizes many of the ideas first described by Milward and Beverdige (Milward and Beveridge 2003, Milward and Beveridge 2004), using modern Semantic Web frameworks. Let us first summarize OntVPA's distinguished features and architecture so that we can describe the DSL improvements "on top" of the OntoVPA model.

### The Original OntoVPA Model & Architecture

OntoVPA employs the *Web Ontology Language (OWL)*[3] for knowledge representation and reasoning (Baader et. al. 2003). Additionally, it uses ontology-based SPARQL[4] rules for DM which are executed in a custom VPA-specific rule engine built on top of the JENA reasoner[5]. In what follows, we assume basic familiarity with OWL notions such as class, instance, and property, and likewise for SPARQL (Baader et. al. 2003). These SPARQL rules are SPARQL queries extended with some annotations: rules have IDs, priorities for conflict resolution, can pass control to other rules, etc. This results in a custom, DM-specific ontology-based highly expressive rule language and engine – the OntoVPA rule engine.

An OntoVPA model has the following main components:
- OWL ontology for *background and domain knowledge* – e.g., for the Restaurant Recommendation VPA it will have classes (concepts), relations (OWL object and datatype properties) and instances (individuals) for representing (types of) restaurants, cities, different cuisines, address and recommendations, an actual database of restaurant instances, etc. Frequently, we extent and reuse Schema.org[6].
- OWL ontology of *classes and relations for dialogue / discourse representation* – this includes classes for user and system turns, intents and their slot values, and system response classes. *Speech act theory* (Koller and Searle 1969) provides us with a coarse but useful upper ontology, for Requests, Response, etc.; also compare (Fernández-Rodicio et. al. 2020) for a similar upper-level ontology. *At runtime,* these classes and relations are instantiated in an *OWL ABox, which is a set of class instances and relationships,* representing the actual dialogue, i.e., the history of user and system dialogue turns. For example, a restaurant recommendation VPA will have a user intent class `FindRestaurantIntent`, and a corresponding system response class `SuggestRestaurantResponse`, and relations (object properties) for "slots" such as `cuisine` and `location`.
- A layer of *Generic Dialogue Management Rules* – these rules implement core DM capabilities, like slot-value filling, anaphora resolution, context-based disambiguation, details of turn taking, etc. Not every domain requires all these capabilities; they can be disabled individually, but also refined as needed.
- VPA *domain-specific rules that implement the dialogue workflow and intent processing*. In a Restaurant Recommendation VPA, there will be rules that trigger if certain intents are instantiated, e.g., a `RecommendRestaurant` rule triggers if the current intent is a `FindRestaurantIntent`. The rule checks for the presence of the requested `location` and `cuisine`, then employs ontology-based query answering to retrieve matching restaurant instances and presents the result.

We heavily rely on *upper-level ontologies and inheritance,* and on generic and reusable behavior specified in the generic DM rules. The combination of ontology reasoning with expressive, succinct higher-order ontology-based rules is the defining feature of OntoVPA.

## Ontology-Based Dialogue Management Example

Let us illustrate *the interplay of ontology reasoning, generic DM rules, and domain-specific rules that implement the dialogue workflow* by means of the following dialogue in the Restaurant Recommendation Domain:

- *User*: I am looking for a restaurant!
- *VPA*: In what city?
- *User*: In Palo Alto.
- *VPA*: How about McDonalds?
- *User*: Chinese please.
- *VPA*: Got it – Su Hong on 4256 El Camino Real?

The first utterance can clearly be classified as a `FindRestaurantIntent` by the statistical intent classifier - let us assume this intent class has a required `slot location` (of range `City`), and an optional slot `cuisine` (of range `Cuisine`). This intent is instantiated in the dialogue representation ABox, but without any slot values. A generic DM rule now inspects the current intent and the definition of its associated OWL intent class and determines that a required slot value (the `location` slot) is missing. To check for the presence or absence of slot values, this SPARQL rule uses *existential quantification over slots / predicates*; consequently, we consider this a *higher-order rule*. It hence triggers a follow-up question to the user, who subsequently answered "Palo Alto".

The string "PaloAlto" is automatically mapped to the corresponding OWL `City` instance of the same name. Next, another generic DM rule determines that the `PaloAlto` individual is substitutable for the inquired `location` slot value of the previous intent - the user has answered the question. Hence, `PaloAlto` is filled in as a slot value of the previous `FindRestaurantIntent`, and the intent is now marked *completely specified* by another DM rule. This subsequently triggers the domain-specific workflow `RecommendRestaurant` rule, which proceeds as already described. It retrieves and presents a single restaurant to the user, who subsequently requests "Chinese" instead of the presented option.

Since no intent was recognized from the word "Chinese" in isolation, the system relies again on the dialogue history for sense making / understanding in context. The ontology-based parser has mapped "Chinese" to the class `ChineseCuisine`, which is a subclass of `Cuisine`. From the dialogue history it determines that `ChineseCuisine` is a potential slot-filler for the `cuisine` slot on the previous `FindRestaurantIntent`, and it is hence augmented with the optional slot filler. Given that the intent has changed, it is then re-executed by a rule that detects the change and which then triggers re-execution of the `RecommendRestaurant` rule.

Many of OntoVPA's generic DM rules are specified in a similar flavor. Central expressive means of these higher-order generic SPARQL rules are the ability to

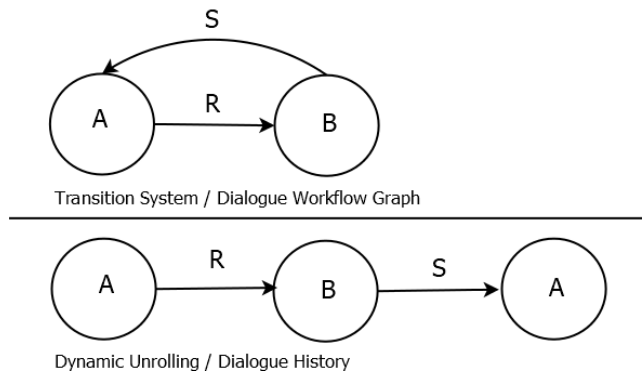

Figure 1 – Illustration Transition System & History Unrolling

- introspect the OWL class definitions (of intents),
- traverse and assess the full dialogue history,
- perform ontology reasoning (e.g., sub-class and sub-property inferences) in the rule itself,
- existentially and universally quantify "in a second-order fashion" over arbitrary classes and slots (properties) rather than having to codify individual rules for individual classes and slots,
- create and update arbitrarily nested and complex structured graph representations (in the ABox); SPARQL allows for the creation of Skolem instances in the right-hand side of rules, and of arbitrary atoms / structure,
- specify defeasibility and priorities on rules.

These DM rules are supplied in an upper level, reusable rule layer applicable to any domain - they provide generic DM capabilities. In addition, OntoVPA modelers typically still must write domain specific rules like the `RecommendRestaurant` rule, which is the highest LOE activity that we are seeking to replace with the DSL here.

## Modeling Dialogue Workflows in OntoVPA

Our underlying *conceptual model* is the

- *Dialogue Workflow Graph (DWG).* Like a *(labeled) transition system*[7], this graph represents the space of possible dialogues and workflows, and hence the policy of the system.

This is a development time (static) representation. At runtime, this conceptual graph is interpreted and operationalized by the Dialogue Manager to compute the system answers / turns. This results in the dynamic

- *Dialogue (Discourse) History*, which is an OWL ABox in OntoVPA. OWL instances are representing user and system turns (nodes), which are asserted and computed, with their causal, temporal, and thematic attributes and relationships.

The dialogue history can be considered an unrolling of the conceptual DWG, analog to the unrolling of a transition system into a trace of dynamic system behavior, see Figure 1.

```
CONSTRUCT {
    _:s vpa:assertedType vpa:B .
    _:s vpa:message "Transitioned to B!" .
    _:s vpa:assertedType vpa:CurrentSystemStepMarker .
    ?i vpa:computedSystemResponse _:s }
WHERE
 { ?i vpa:assertedType vpa:CurrentUserStepMarker .
   ?i vpa:assertedType vpa:R .
   ?i vpa:previousUserStep ?p .
   ?p vpa:computedSystemResponse ?pr .
   ?pr vpa:assertedType vpa:A }
```

Figure 2 – Simple SPARQL Transition Rule

In this abstract example, *A* might represent a `Confirm-RestaurantLocation` state, *R* a corresponding user utterance ("In Palo Alto!"), and *B* a `SuggestRestaurant` system response.

The dialogue history employs a notion of *current user turn* and *current (last) system turn*. User turns are simply asserted in the dialogue history (and made current) by the intent classifier and ontology-based slot-filler, whereas the system turns are computed by the Dialogue Manager.

In the following, we describe three *concrete* representation options for the *conceptual* DWG, corresponding *to layers of representations* in OntoVPA. Level 1 is the original OntoVPA representation, Level 3 the new DSL, and Level 2 an intermediate representation.

## Level 1 - Rule-Based Workflow Graph Modeling

It is straightforward to operationalize a conceptual DWG, or transition system, via rules:

> For each labeled edge $A \rightarrow^R B$ in the DWG, create a corresponding rule that, given current state *A*, creates a new successor node of state *B* if condition *R* holds.

A corresponding OntoVPA SPARQL rule is shown in Figure 2. If, in the `WHERE` antecedence, the current user step satisfies *R* and the previous system response was of type *A,* then the `CONSTRUCT` consequence of the SPARQL rule creates a new node of type *B* via the _:s Skolem constant constructor. The freshly constructed node will be asserted as current system step into the dialogue history, and the asserted `message` property will cause a system utterance. Note that *A, B, R,* are OWL ontology classes and properties, and that the SPARQL engine is aware of the deductive closure of the OWL ontology (i.e., its inferences).

The type and number of *conditions* (*R* in the example) can be arbitrarily complex and may involve full OWL / Description Logic reasoning over the current dialogue, workflow completion, and world state.

There is no limit on what can be asserted with a `CONSTRUCT` consequent of a SPARQL rule at dialogue runtime – not only can we update the dialogue history, but also other parts of the ABox. For example, personal information learned about the user could be stored on the user's profile information in the ABox.

Two obvious drawbacks of the just discussed rule-based modeling style are: a) the large number of rules required, and b) the amount of boilerplate code needed (e.g., each SPARQL rule must determine the `CurrentUserStep` in the history, construct the follow-up node, mark it as `CurrentSystemStep`, etc.) Regarding a), the number of rules is roughly given by

$$\#nodes * average\_node\_out\_degree$$

This number *is quickly in the three-figure range* for non-trivial dialogue models (see below). Problem b) can be alleviated by using our SPARQL macros. However, the "graph as a set of rules" representation is also unwieldy from a modeling perspective, and a direct graph representation would be preferable. Moreover, SPARQL modeling is intellectually demanding and requires expert knowledge.

## Level 2 - Graph-Based Workflow Graph Modeling

Our first step of remediation is hence to represent the workflow graph *directly as a graph* in the OWL ABox, *in terms of instances (nodes) and relations (edges)*.

Unlike a set of SPARQL rules, an ABox graph representation is not an executable specification - hence, a *workflow graph interpreter* is required that operationalizes the graph. We chose to implement the interpreter (in a meta-circular way) *itself* in terms of OntoVPA's SPARQL rules, instead of extending the JENA-based rule engine.

The ABox-based, explicit and direct graph representation has some modeling benefits – SPARQL domain rules no longer need to be authored by the domain developer. The interpreter rules are part of the OntoVPA framework already and usually do not need to be altered, just like the generic DM rules already discussed. Moreover, DWGs can now be inspected and, in principle, also edited visually, by means of OWL tools such as Protégé[8] (Musen 2005).

One drawback of this representation though is that it requires *additional modeling vocabulary*, e.g., classes and relationships for representing logical conditions and control structure, as we can no longer rely on SPARQL specifications. The representation is *considerably less succinct* - a graph of hundred nodes might require a few thousand ABox assertions (see below for concrete numbers). The direct representation hence suffers from a *boilerplate representation problem* and complicates modeling due to a lack of tool support for our extra- and control vocabulary.

Hence, a final abstraction layer added on top, Level 3. This layer provides the high-level DSL which gets compiled in the Level 2 representation just discussed. We can consider Level 2 as semantic virtual machine code produced from Level 3 DSL high-level specifications, whereas Level 1 provides the generic implementation of the virtual machine that implements the DM / VPA instruction set in OntoVPA.

```
(defnode n1
  (:condition A)  ; node condition
  (:message "In node n1") ; output
  (:transition ; edge condition
    (R n2))) ; transition to n2 if R
(defnode n2
  (:condition B)
  (:message "Transitioned to n2"))
```

Figure 3 – DSL-Based Transition Specification

```
(defnode node_mhc2
  (:message "Where is the bleeding?")
  (:transition
    (Limb       node_limb)
    (HeadOrNeck node_head_or_neck))
  (:extract-and-store
    (BodyPart currentUser hemorrhageLocation)))
```

Figure 4 – Updating the ABox with Extract-And-Store

## Level 3 - DSL-Based Workflow Graph Modeling

In our DSL, a workflow graph node is described together with its outgoing transitions and conditions, see Figure 3. Here, a transition from node *n1* (satisfying *A*) to node *n2* (satisfying *B*) if *R* holds, is specified.

Transitions are activated by *edge conditions*, and in addition, there is a *notion of active and disabled nodes*, based on the truth values of their corresponding *node conditions*. The interpreter will never transition into a disabled node.

A variety of *condition types* (specified via different *clauses* that start with a colon) exist. Node and transition conditions can be specified based on:

- the *current intent and its slot values*. This information is usually asserted by a statistical intent classifier and the ontology-based slot filler. However, training statistical intent classifiers can be costly, and we hence support building VPAs without them, by offering
- parsing raw textual user input (e.g., from ASR) via *ontology-based grammar expressions*. The domain ontology then also plays the role of a *lexicon or thesaurus*, providing domain terms, and its hyponyms, hypernyms, synonyms, and antonyms. We support *regular expressions over ontology terms* - for example, the expression (Neg Ampl PosDesc) is satisfied by the utterance "not very good", or "not so well", and might be used to transition to some other node. Sometimes, simple ontology-based keyword spotting is sufficient as well. We usually accept a hyponym for the specified ontology term in the expression, but this is not a strict requirement.
- complex (negated) *logical conditions that are evaluated over the ABox*, having access to the dialogue history and (ABox) domain model. Logical conditions are specified as *path expressions of OWL properties and classes,* starting from either an ABox individual (that then has to satisfy the expression), or from a class name (then there needs to be some instance that satisfies the expression). For example, the path expression ((:ind currentUser)(healthCondition Cough)) evaluates to true if the currentUser individual has a healthCondition (object property) slot value that satisfies (i.e., is an instance or a subclass of) the Cough class.

It is not always desirable or feasible to anticipate all potential dialogue transitions in terms of outgoing conditioned edges on nodes at development time. To allow a more naturally flowing and less linear dialogue, we allow transitions into different dialogue graph regions, e.g., corresponding to different conversation topics, *without requiring explicit outgoing edges on nodes leading into these regions.* To facilitate these transitions into different regions, we employ the notion of a *trigger* for a node. If the trigger on a node is satisfied, control can transition into the trigger-activated node *in a non-local way, i.e., no edge needs to be transitioned to arrive at it.*

Additional annotations are specified on the node-level, determining aspects of control and thus direct the interpreter

- whether the node is a *start or end node* of a "topic", i.e., a *sub-graph* that initiates a dialogue / conversation about a specific topic,
- if it *can be triggered*, and hence gain control, if a certain *trigger condition* is satisfied (i.e., an intent or special event has been recognized and been asserted as current),
- whether the node allows to relinquish control to another node via trigger-based / non-local transitions,
- whether the dialogue should return and resume at that point after the node lost control in a non-local way (i.e., after the triggered "sub-dialogue" has completed),
- if, in case of a topic end node, the control should *return* to the previously active node, or simply continue;
- if the node is *modal*, i.e., is waiting for input before continuing / transitioning to the next node, or if the transition is *immediate* (these nodes are useful for breaking up the workflow into smaller chunks, for example).

Finally, the last group two groups of node annotations allow us to a) *generate output* and b), to *update the ABox.*

For a), each node can be annotated with a multitude of messages using the :message field. We support template-based NLG – rather than using variables for the template "holes", we are yet again using ontology-based path expressions that "peek into the ABox" to fill the template holes.

For b), we can assert and retract arbitrary ABox assertions. An example from a OntoVPA Medical Decision Support System is shown in Figure 4: the medic is asked for the location of a bleeding. Any answer subsumed by the BodyPart class is then extracted from the utterance and stored as a slot value on the currentUser individual's hemmorhageLocation slot, using the extract-and-store clause. Depending on the specific BodyPart, the system then transitions into specific branches for hemorrhage treatment (for Limb or HeadOrNeck).

The *DSL Compiler* is implemented in Racer (Haarslev et. al. 2012), and a visual representation is generated as a by-product using DOT and Graphviz[9].

## Quantifying the DSL Benefits

| Domain | #Nodes | #Rules saved | #Assertions | RpN | ApN |
|---|---|---|---|---|---|
| ChatPal | 109 | 291 | 2520 | 2.7 | 23 |
| Medic | 29 | 26 | 374 | 0.9 | 11 |

Table 1- Quantifying the DSL Benefits

Table 1 shows the *number of nodes, number of rules that did not have to be modeled, number of ABox graph assertion generated by the DSL compiler, the average number of Rules per Nodes (RpN), and the number of Assertion per Node (ApN)* for 2 domains, ChatPal and Medic:

- *ChatPal*: A virtual personal companion for the elderly that aims to overcome loneliness and social isolation. ChatPal flexibly engages the senior in templated conversations about specific topics, such as hobbies, the senior's life history, family, relationships, etc., and it learns about the users' interests, hobbies, and relationships and is able to leverage the learned knowledge in subsequent sessions.
- *Medic*: A simple Medical Decision Support System prototype. It implements the workflows from a standard medical procedure handbook. Using the DSL, its workflow dialogue graph was developed in ~20 hours only.

From our experience, each Level 1 OntoVPA rule modeled by an OntoVPA expert accounts for ~1.5 hours LOE. In contrast, Medic was modeled using DSL with 0.68 hours LOE per DSL node; modeled on Level 1, the corresponding 26 rules would have required 26 * 1.5 = 39 hours - a reduction by 48.7 %. For ChatPal, the situation is even more significant: 1.5 * 291 = 436.5 vs. 109 * 0.68 = 74.12 hours – a reduction by 83 %. For illustration, the generated visual DWGs are given for ChatPal[10] and Medic[11] in the endnotes.

## Related Work

Knowledge bases and ontologies have been used since the early days of the LUNAR system (Woods and Kaplan 1977) for Natural Language Understanding, Question Answering and Dialogue Systems, and, more recently, in HALO (Gunning et. al. 2010). Frequently, (OWL) ontologies are being used by the Dialogue Manager for ontology-driven question answering relying on domain ontologies and external knowledge sources (Pérez et. al. 2006), (Wantroba and Romero 2014). Other case studies focused on ontology modeling and dialogue design based on task structures represented in OWL (Chaudhri et. al. 2006). OntoDM (Altinok 2018) uses OWL ontologies for domain representation and partially for the dialogue history and NLU tasks such as anaphora resolution (Milward and Beveridge 2003), but unlike OntoVPA it relies on special purpose algorithms that are informed by the ontologies, and is thus not fully declarative. OWLSpeak (Heinroth et. al. 2010, Ultes et. al. 2016) is based on Information State Theory (Larsson and Traum 2000), but the state is not implemented in OWL at runtime; hence, no ontology-based rules are being used. VOnDA (Kiefer et. al. 2019) is similar to OntoVPA in that it employs OWL and uses a proprietary RDFs/OWL-inference aware "reactive rules" language for policy / dialogue flow specification, whereas OntoVPA relies on extended SPARQL over an ABox. The Converness system (Meditskos et. al. 2020) is a hybrid system that uses OWL2 for most aspects of dialogue and domain representation, SPARQL endpoints for question answering, and defeasibility rules for context-aware reasoning (Mavropoulos et. al. 2019). The rules support disambiguation and conflict resolution over the dialogue discourse. Stoyanchev and Johnston (Stoyanchev and Johnston 2018) implement the Information State Approach (Larsson and Traum 2000) using Knowledge Graphs, also drawing inspiration from OntoVPA.

An early approach to a generic, visual dialogue flow specification language is given in (Kölzer 1999): a compiler translates visual dialogue flow graphs into Prolog knowledge bases. The dozens of contemporary commercial visual flow builders for chat bot development (Google's Dialogflow, bots, Visual Chatbot Builder, etc.) usually require some form of programming for customization once the boundaries of the narrowly defined standard domains are reached (e.g., pre-defined models are supplied that don't generalize well). We believe that OntoVPA's DSL provides a more general and more productive modeling environment.

## Conclusion

We have made steps towards improving dialogue modeling efficiency and customer participation for ontology-based VPAs with our new dialogue modeling language, DSL. It corresponds to a new abstraction layer on top of OntoVPA, comes with a less technical DWG modeling syntax, a compiler and visualizer, and was successfully applied in 2 projects, reducing dialogue modeling effort significantly (by an estimated 49 % and 83 %, respectively). Our solution is not based on visual programming or modeling – we believe that the throughput of textual modeling languages is higher and leads to more general solutions. Future work will focus on extending tool support, potentially making the framework available to other users, and applying it in more domains.

**Acknowledgements**
This work is supported by the US Army Medical Research and Materiel Command under Contract No. W81XWH-19-C-0096. The views, opinions and/or findings contained in this report are those of the author(s) and should not be construed as an official Department of the Army position, policy or decision unless so designated by other documentation. We would also like to thank Karen Myers for feedback and suggestions that significantly helped to improve the paper.

---

[1] Gartner prediction: https://www.gartner.com/en/newsroom/press-releases/2019-01-09-gartner-predicts-25-percent-of-digital-workers-will-u

[2] Kasisto Company Website: www.kasisto.com

[3] OWL 2 Specification: www.w3.org/TR/owl-overview/

[4] SPARQL Specification: www.w3.org/TR/sparql11-query/

[5] Apache Jena Website: jena.apache.org

[6] Schema.org Website: schema.org

[7] For example: en.wikipedia.org/wiki/Transition_system

[8] Protégé project: protege.stanford.edu

[9] DOT and Graphviz: https://en.wikipedia.org/wiki/DOT_(graph_description_language)

[10] ChatPal generated DWG: www.ai.sri.com/~wessel/chatpal.pdf

[11] Medic generated DWG: www.ai.sri.com/~wessel/medic.pdf